\documentclass[letterpaper]{article} 

\usepackage{aaai2026}
\nocopyright

\usepackage{times}  
\usepackage{helvet}  
\usepackage{courier}  
\usepackage[hyphens]{url}  
\usepackage{graphicx} 
\usepackage{natbib}  
\usepackage{caption} 
\usepackage{algorithm}
\usepackage{algorithmic}

\usepackage{amsmath,amssymb}
\usepackage{makecell}
\usepackage{multirow}
\usepackage{booktabs}
\usepackage{bm}
\usepackage{amsfonts}
\usepackage{booktabs}
\usepackage{subcaption}
\usepackage{graphicx}

\usepackage{pifont}
\newcommand{\cmark}{\ding{51}}  
\newcommand{\xmark}{\ding{55}}  
\newcommand{\ms}[2]{{#1\tiny{$\pm$#2}}}
\usepackage{verbatim}

\makeatletter
\def\ul#1#2#3{\underline{#1{#2}{#3}}}
\makeatother

\usepackage{xspace}
\makeatletter
\DeclareRobustCommand\onedot{\futurelet\@let@token\@onedot}
\def\@onedot{\ifx\@let@token.\else.\null\fi\xspace}
\def\eg{\emph{e.g}\onedot} 
\def\ie{\emph{i.e}\onedot} 
 
 \def\vs{\emph{vs}\onedot}

\makeatother

\usepackage{newfloat}
\usepackage{listings}
\DeclareCaptionStyle{ruled}{labelfont=normalfont,labelsep=colon,strut=off} 
\floatstyle{ruled}
\newfloat{listing}{tb}{lst}{}
\floatname{listing}{Listing}
\title{DGP: A Dual-Granularity Prompting Framework for Fraud Detection with Graph-Enhanced LLMs}

\author {
    Yuan Li\textsuperscript{\rm 1},
    Jun Hu\textsuperscript{\rm 1},
    Bryan Hooi\textsuperscript{\rm 1},
    Bingsheng He\textsuperscript{\rm 1},
    Cheng Chen\textsuperscript{\rm 2}
}
\affiliations {
    \textsuperscript{\rm 1}National University of Singapore
    \textsuperscript{\rm 2}ByteDance Inc.\\
    li.yuan@u.nus.edu, \{jun.hu, dcsbhk, dcsheb\}@nus.edu.sg, chencheng.sg@bytedance.com
}

\usepackage{bibentry}

\begin{document}

\maketitle

\begin{abstract}
Real-world fraud detection applications benefits from graph learning techniques that jointly exploit node features—often rich in textual data—and graph structural information.
Recently, Graph-Enhanced LLMs emerge as a promising graph learning approach that converts graph information into prompts, exploiting LLMs' ability to reason over both textual and structural information.
Among them, text-only prompting, which converts graph information to prompts consisting solely of text tokens, offers a solution that relies only on LLM tuning without requiring additional graph-specific encoders.
However, text-only prompting struggles on heterogeneous fraud-detection graphs: multi-hop relations expand exponentially with each additional hop, leading to rapidly growing neighborhoods associated with dense textual information. These neighborhoods may overwhelm the model with long, irrelevant content in the prompt and suppress key signals from the target node, thereby degrading performance.
To address this challenge, we propose Dual Granularity Prompting (DGP), which mitigates information overload by preserving fine-grained textual details for the target node while summarizing neighbor information into coarse-grained text prompts.
DGP introduces tailored summarization strategies for different data modalities---bi-level semantic abstraction for textual fields and statistical aggregation for numerical features---enabling effective compression of verbose neighbor content into concise, informative prompts. 
Experiments across public and industrial datasets demonstrate that DGP operates within a manageable token budget while improving fraud detection performance by up to 6.8\% (AUPRC) over state-of-the-art methods, showing the potential of Graph-Enhanced LLMs for fraud detection.
\end{abstract}

\section{Introduction}

\begin{figure}[!t]
\centering
\begin{subfigure}{\linewidth}
    \centering
    \includegraphics[width=\linewidth]{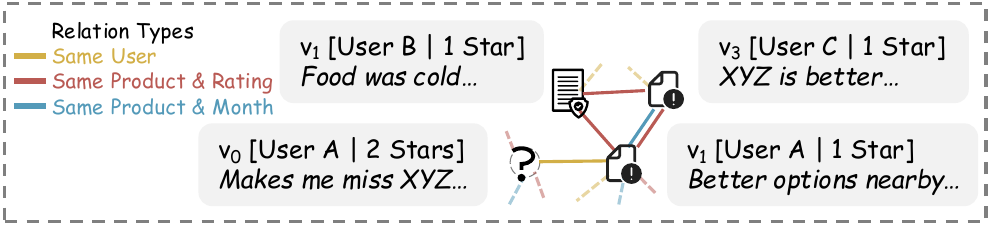}
    \caption{Fraudsters exhibit rich semantic patterns}
    \label{fig:comp-graph}
\end{subfigure}

\vspace{0.5em}

\begin{subfigure}{\linewidth}
    \centering
    \includegraphics[width=\linewidth]{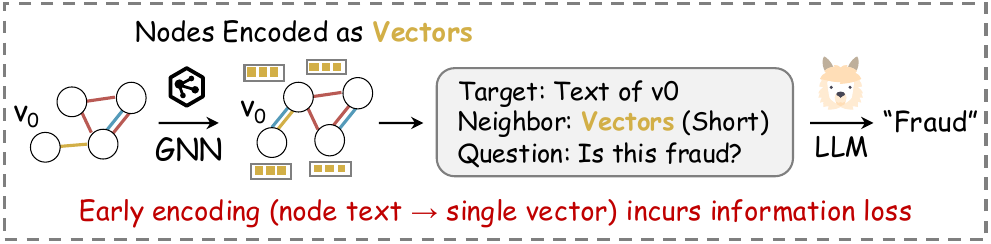}
    \caption{Encoding-based prompting}
    \label{fig:comp-encoding}
\end{subfigure}

\vspace{0.5em}

\begin{subfigure}{\linewidth}
    \centering
    \includegraphics[width=\linewidth]{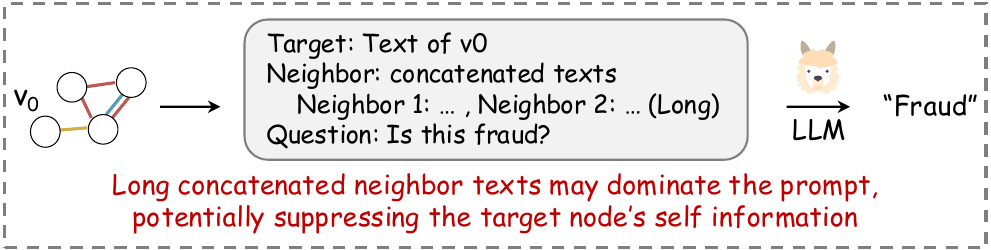}
    \caption{Text-only prompting}
    \label{fig:comp-text}
\end{subfigure}

\caption{Graph-to-prompt methods for fraud detection.}
\label{fig:comparison}
\end{figure}

Graph-based fraud detection has emerged as a critical research direction, driven by its effectiveness in capturing the complex relational patterns inherent in real-world data~\cite{xu2024revisiting,akoglu2015graph,rayana2015collective}.
The intricate structural properties of graphs, combined with the rich semantic and numerical information on nodes, present unique opportunities and challenges for effectively identifying fraudulent entities. Real-world applications such as anomaly detection in social networks~\cite{chen2024consistency,sharma2018nhad}, fake account identification~\cite{li2022sybilflyover,hooi2017graph}, and the detection of malicious user-generated content~\cite{rayana2015collective,mcauley2013amateurs} benefit from advanced graph learning techniques.

\textbf{Graph-Enhanced LLMs for Fraud Detection.} 
In recent years, various Graph Neural Networks (GNNs) have been proposed for graph-based fraud detection, achieving notable success by leveraging neighborhood information and structural patterns to enhance detection accuracy~\cite{duan2024dga,li2024sefraud}. 
More recently, graph-enhanced Large Language Models (LLMs) have emerged as a promising alternative for graph-based fraud detection tasks, leveraging their generalizable language capabilities and demonstrating competitive performance across a range of tasks~\cite{tang2024graphgpt,tang2024higpt,liu2024anomalyllm}. 
These approaches have shown potential in analyzing the rich semantics associated with fraudulent nodes, as well as the diverse relationships among them (as illustrated in Figure~\ref{fig:comp-graph}), by exploiting the semantic nuances within the graph~\cite{tang2024graphgpt}. Notably, we distinguish these methods from LLM-enhanced GNNs such as TAPE~\cite{he2024harnessing} and FLAG~\cite{yang2025flag}, which incorporate LLM-encoded features and rely heavily on the classification capabilities of GNNs. 
In this work, we focus on leveraging graph-enhanced LLMs as standalone classifiers to fully explore their potential in graph-based fraud detection.

To bridge the gap between graph-structured data and LLMs, graph-enhanced LLMs transform graph data into textual prompts (graph-to-prompt) to naturally integrate both graph structure and semantics into LLMs~\cite{fatemi2023talk,ye2023language}.
Two major graph-to-prompt strategies, as depicted in Figure~\ref{fig:comp-encoding} and~\ref{fig:comp-text}, have been developed in recent literature: (1) Encoding-based prompting, exemplified by approaches such as GraphGPT~\cite{tang2024graphgpt} and HiGPT~\cite{tang2024higpt}, encodes nodes into compact vectors and subsequently feeds them into an LLM. These methods substantially reduce prompt length via node encoding, but suffer from early vectorization, leading to \textbf{information loss} due to reduced semantic-level interactions~\cite{li2023survey}. In contrast, (2) text-only prompting~\cite{wang2023can,
ye2023language,fatemi2023talk,zhu2025llm} preserves detailed semantic interactions by concatenating neighbor texts into the prompt.
However, these methods inherently suffer from excessive prompt length, leading to \textbf{distraction from crucial content} due to information overload.
For example, in industrial scenarios, each neighboring node can be associated with over 1,500 tokens, resulting in a 2-hop neighborhood with up to 2 million tokens, which poses challenges for incorporating dense textual information for fraud detection.

\begin{figure}[!t]
  \centering
  \begin{subfigure}[b]{0.49\linewidth}
    \centering
    \includegraphics[width=\linewidth]{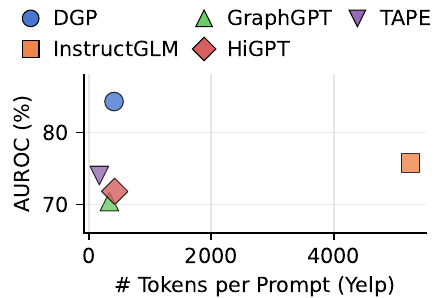}
  \end{subfigure}
  \hfill
  \begin{subfigure}[b]{0.49\linewidth}
    \centering
    \includegraphics[width=\linewidth]{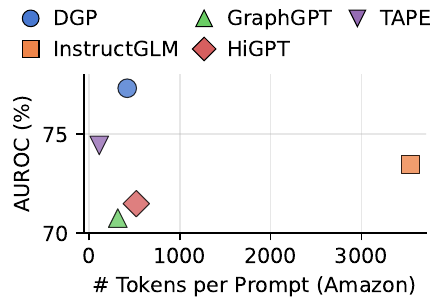}
  \end{subfigure}
\caption{Fraud detection performance ($\uparrow$) \vs token usage per prompt ($\downarrow$) across different methods and datasets. Our proposed method, DGP, achieves top performance with moderate token consumption, demonstrating a notable balance between token usage and performance.}
  \label{fig:tok-perf}
\end{figure}

In this work, we propose \textbf{D}ual \textbf{G}ranularity \textbf{P}rompting (DGP), a novel text-only prompting framework that leverages the rich semantics on graphs while addressing the challenge of excessive prompt length. To reduce the information loss incurred by early-stage encoding, DGP selectively preserves fine-grained text for the target node while summarizing neighbors retrieved from different metapaths into compact, coarse-grained texts. For textual features, we employ bi-level semantic summarization to reduce the prompt length. For numerical features, we adopt precise numerical summarization to retain key insights.
As illustrated in Figure~\ref{fig:tok-perf}, our approach achieves an impressive balance between token usage and performance. Compared to prior state-of-the-art methods, DGP operates with a manageable prompt length while improving fraud detection performance by up to 6.8\% (AUPRC), demonstrating the effectiveness of our dual-granularity design with reasonable token budgets.

The key contribution of this work is three-fold:
\begin{itemize}
\item We propose DGP, a novel graph prompting framework that integrates fine-grained textual details for target nodes with coarse-grained semantic summaries for their neighbors, thereby overcoming limitations faced by existing graph-to-prompt methods.
\item We introduce specialized summarization strategies for compressing neighborhoods associated with textual and numerical features into concise, semantically meaningful prompts tailored for LLM processing.
\item Extensive experiments on public and industry datasets demonstrate the superior empirical performance of DGP, achieving manageable prompt lengths while improving fraud detection performance by up to 6.8\% in AUPRC compared to state-of-the-art approaches.
\end{itemize}

\section{Related Work}

\subsection{Graph Neural Networks for Fraud Detection}
Graph neural networks (GNNs) have become the dominant approach for fraud detection by modeling relational patterns in graphs~\cite{akoglu2015graph,rayana2015collective,duan2024dga,li2024sefraud}. Classic models such as GCN~\cite{kipf2017semisupervised} and GAT~\cite{veličković2018graph} have inspired many variants targeting specific challenges, including camouflage (CARE-GNN~\cite{dou2020enhancing}), heterophily (PMP~\cite{zhuo2024partitioning}), and limited supervision (ConsisGAD~\cite{chen2024consistency}, barely-supervised learning~\cite{yu2024barely}). 
However, most GNN-based approaches underutilize the fine-grained textual semantics widely available in real-world graphs, which our method explicitly addresses.

\begin{figure*}[!t]
\centering
\includegraphics[width=\linewidth]{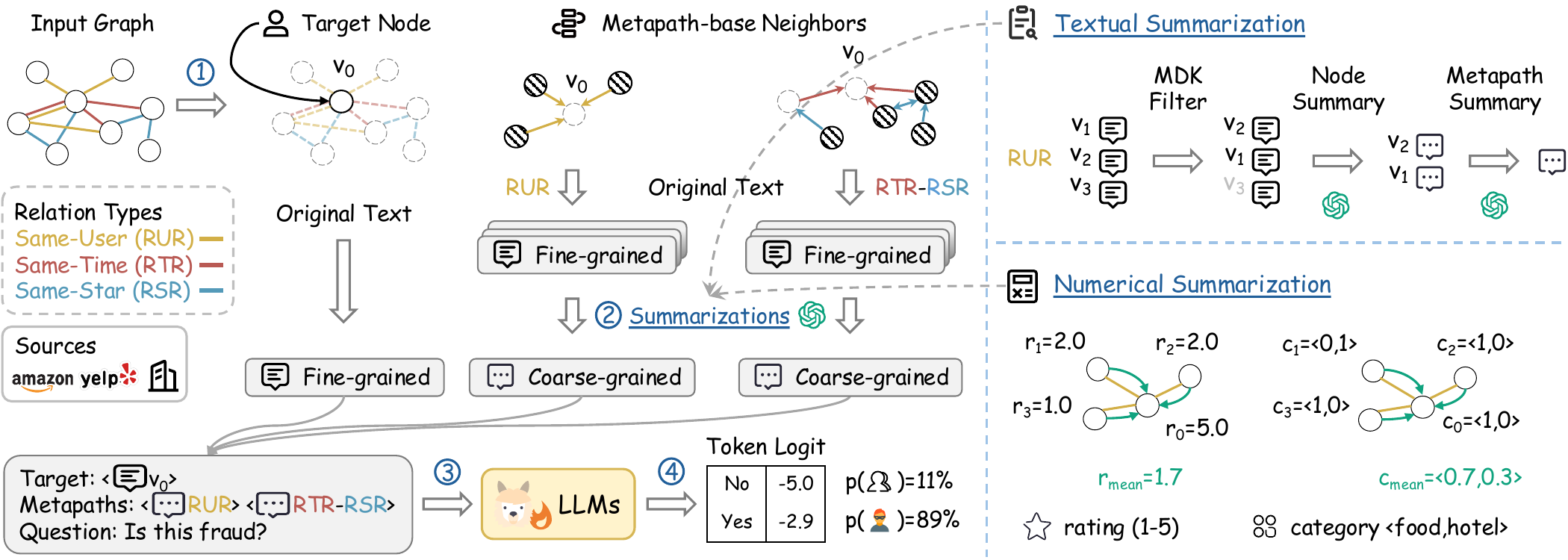}
\caption{Overview of the proposed DGP framework.}
\label{fig:framework}
\end{figure*}

\subsection{Integrating LLMs with Graphs}

Recent advances in integrating LLMs with graph data can be broadly classified into \textit{graph-enhanced LLMs} and \textit{LLM-enhanced GNNs}.
Graph-enhanced LLMs primarily adopt graph-to-prompt strategies, which can be divided into encoding-based prompting and text-only prompting. Encoding-based prompting~\cite{tang2024graphgpt,tang2024higpt} compresses graph features for LLM input, potentially resulting in semantic loss.
Specifically, GraphGPT~\cite{tang2024graphgpt} aligns LLMs with graph structural information via a dual‐stage instruction‐tuning paradigm and a graph‐text alignment projector.
HiGPT~\cite{tang2024higpt} extends instruction tuning to heterogeneous graphs by introducing an in‐context heterogeneous‐graph tokenizer and heterogeneity‐aware fine‐tuning.
In contrast, text-only prompting~\cite{fatemi2023talk,ye2023language,zhu2025llm} concatenates the texts of neighboring nodes as input to LLMs, which may lead to excessively long prompts and distract from crucial information.
For example, InstructGLM~\cite{ye2023language} frames graph tasks as natural-language instructions for generative LLMs, enabling node classification on citation networks.

Another line of work, LLM-enhanced GNNs, integrates LLM-encoded features into GNNs to improve node representation. For example, TAPE~\cite{he2024harnessing} uses LLM-generated explanations as auxiliary features for downstream GNNs, and FLAG~\cite{yang2025flag} leverages discriminative text extraction to address neighborhood camouflage in fraud detection.
While these approaches rely on GNNs as the primary modeling framework and treat LLMs as feature encoders, our method falls under the graph-enhanced LLMs category, where LLMs serve as the core classifier and directly operate on graph-structured information. This distinction allows us to fully exploit the capabilities of LLMs for graph-based fraud detection.

\section{Preliminaries}

In this section, we formalize the fraud detection problem on heterogeneous graphs and define metapaths.

\subsection{Graph-based Fraud Detection}\label{subsec:problem}
Given a heterogeneous graph
\(
G=\{V,E,\mathcal{R},\mathcal{X}\}
\),
where \(V\) denotes a set of $N$ nodes,  
\(E\subseteq V\times V\times\mathcal{R}\) is the set of typed edges,  
\(\mathcal{R}=\{r_{1},\dots,r_{|\mathcal R|}\}\) is the set of edge relation types, and $\mathcal{X} = \{x_{v}\}_{v\in V}$ represents the node features of mixed types.
Each node \(v\in V\) is associated with a feature tuple  
\(
x_{v}=(x^{\text{text}}_{v},\,x^{\text{num}}_{v})
\),
where \(x^{\text{text}}_{v}\) denotes raw textual content (\eg, user-written reviews) and \(x^{\text{num}}_{v}\in\mathbb{R}^{d}\) stacks all numeric or one-hot categorical features (\eg, a rating ranging from 1 to 5 stars).

We focus on the task of node-level fraud detection. Let \(y_{v}\in\{0,1\}\) be a binary label indicating whether node \(v\) is fraudulent (\(1\)) or benign (\(0\)).  
The objective is to learn a function $f: V \longrightarrow \{0,1\}$ that minimizes the empirical risk:
\begin{equation}
\mathcal{L}=\frac1{|V_{\text{train}}|}\sum_{v\in V_{\text{train}}}\ell\!\bigl(f(v),y_{v}\bigr)    
\end{equation}
where \(\ell(\cdot,\cdot)\) is the binary cross-entropy loss and \(V_{\text{train}}\subset V\) denotes the labeled training nodes.  
At inference time, \(f\) is applied to each unseen node to predict its label.

\subsection{Metapaths on Heterogeneous Graphs}\label{subsec:metapath}
For each relation \(r\in\mathcal R\), we define \(A_{r}\in\{0,1\}^{N\times N}\) as the typed adjacency matrix, where entry \((A_{r})_{uv}=1\) if \((u,v,r)\in E\).  
A metapath~\cite{Sun2011PathSim} is a finite sequence of relations:
\begin{equation}
P = r_{1}\!\circ r_{2}\!\circ\cdots\!\circ r_{L}    
\end{equation}
which describes a composite semantic, \eg, $Review \rightarrow User \rightarrow Review$.
The metapath-specific adjacency matrix is computed as:
\begin{equation}
A_{P}=A_{r_{1}}A_{r_{2}}\cdots A_{r_{L}}
\end{equation}
and the metapath-specific neighborhood of \(v\) is defined as:
\begin{equation}
\mathcal{N}_{P}(v)=\{u\in V\mid (A_{P})_{vu}>0\}
\end{equation}

\section{Methodology}
\label{sec:method}

This section details the component design of the dual-granularity prompting framework.

\subsection{Dual Granularity Prompting}

Effectively bridging graph-structured data with large language models requires a careful balance between semantic richness and manageable prompt length. Existing approaches tend to fall short: encoding-based prompting compresses neighborhood information at the expense of crucial semantic cues, while text-only prompting preserves detail but quickly overwhelms LLMs with excessive prompt lengths.
Inspired by the selective granularity strategies in recent GNN work such as RpHGNN~\cite{hu2024efficient}, which demonstrates the benefits of retaining fine-grained target node features while abstracting neighborhood context, we propose Dual Granularity Prompting (DGP). DGP preserves fine-grained textual details for the target node and compresses neighbor information into concise, high-level summaries—striking a practical balance between informativeness and token budget.

As depicted in Figure~\ref{fig:framework}, the DGP framework is composed of three core modules: (i) node-level summarization to distill the essence of each node’s raw text, (ii) diffusion-based metapath trimming to select the most structurally and semantically relevant neighbors along each metapath, and (iii) metapath-level summarization to further aggregate both textual and numerical features. The resulting dual-granularity prompts enable LLMs to effectively process complex graph data, capturing essential fraud-related signals.

\subsection{Textual Summarization}

We detail the bi-level textual summarization process.

\subsubsection{Node-level Summarization}

A significant obstacle in leveraging large language models (LLMs) for graph‐based fraud detection is the sheer volume of textual data associated with nodes, particularly when considering multi-hop neighbors. To effectively address this issue, we first condense the textual description of each node into a concise yet representative intrinsic summary. Formally, given the raw textual feature \(x^{\text{text}}_v\) of node \(v\), we generate a summarized text \(s_v\):
\begin{equation}
s_v = \text{Summarize}(x^{\text{text}}_v; B_{\text{node}})
\end{equation}
where \(B_{\text{node}}\) denotes the token budget per node. 
To avoid hand-crafting domain-specific prompts, we adopt a task-agnostic summarization approach that condenses text within a fixed token budget. Although some detail may be lost, the resulting summaries remain effective for downstream metapath extraction and reasoning. In contrast, we empirically observe that task-specific prompts may underperform in the absence of dataset-specific expertise, as they can misguide the model and degrade summarization quality.

\subsubsection{Diffusion‐based Metapath Trimming}

Metapaths capture composite semantics in heterogeneous graphs by connecting nodes through meaningful multi-hop relational sequences. They enable the construction of rich, type-aware neighborhoods that reflect diverse semantic views.  
Although node-level summarization helps reduce redundancy, directly aggregating information from all neighbors across multiple metapaths remains computationally prohibitive and prone to semantic noise.  
To accurately characterize the local fraud-related context of a target node \(v\), we propose a structure- and semantics-aware metapath trimming method guided by the Markov diffusion kernel (MDK)~\cite{fouss2012experimental}.

For each metapath \(P\), we form the row‑stochastic transition
matrix \(\mathbf T_{P}= \mathbf D_{P}^{-1}\mathbf A_{P}\) from the
metapath-specific adjacency matrix \(\mathbf A_{P}\) and degree matrix
\(\mathbf D_{P}=\operatorname{diag}(\mathbf A_{P}\mathbf 1)\).
Averaging the first \(K\) random‑walk powers, \ie, $K$-hops, yields the Markov diffusion
operator:
\begin{equation}
\mathbf Z_{P}(K)=\frac{1}{K}\sum_{k=0}^{K}\mathbf T_{P}^{k}
\end{equation}

Let \(\mathbf{X} \in \mathbb{R}^{n \times d}\) denote the raw node features. We propagate these features via \(\mathbf{Z}_{P}(K)\) to obtain structure-aware semantic embeddings:
\begin{equation}
\mathbf{h}^{(P)}_i(K) = \left[ \mathbf{Z}_{P}(K) \mathbf{X} \right]_{i:}
\end{equation}
where \(\mathbf{h}^{(P)}_i(K)\) is the embedding of node \(i\) under metapath \(P\), corresponding to the \(i\)-th row of the matrix \(\mathbf{Z}_{P}(K) \mathbf{X}\). The joint diffusion distance between nodes \(u\) and \(v\) is:
\begin{equation}
\delta^{(P)}_{K}(u,v)=\bigl\lVert\mathbf h^{(P)}_u(K)-\mathbf h^{(P)}_v(K)\bigr\rVert_2
\end{equation}
which measures how similarly \(u\) and \(v\) diffuse information along metapath \(P\).

To reduce semantic noise and focus on informative context, we retain only the top-\(M\) nearest neighbors of the target node \(v\) based on diffusion distance:
\begin{equation}
\widetilde{\mathcal N}_{P}(v)=
\operatorname*{TopM}_{u\in\mathcal N_{P}(v)}
\!\bigl(-\delta^{(P)}_{K}(u,v)\bigr)    
\end{equation}
This results in a pruned, structure- and semantics-aware neighbor set suitable for downstream fraud detection tasks.

\subsubsection{Metapath Summarization}

We aggregate the node-level summaries of selected neighbors under each metapath \(P\) into a metapath summary \(S_P(v)\):

\begin{equation}
S_P(v) = \text{Summarize}\bigl(\oplus_{u \in \widetilde{\mathcal{N}}_P(v)} s_u;\;B_{\text{meta}}\bigr)
\end{equation}

where \(\oplus\) denotes concatenation and \(B_{\text{meta}}\) is the token budget per metapath summary. This summarization further reduces redundancy by synthesizing a concise and informative metapath-level textual representation.

\subsection{Numerical Summarization}

Unlike textual data, numerical and categorical features often encode precise signals critical for fraud detection. To retain this information, we perform mean aggregation along each metapath. For a given node \(v\) and metapath \(P\), the aggregated representation is defined as:
\begin{equation}
a_P(v) = \frac{1}{|\widetilde{\mathcal{N}}_P(v)|} \sum_{u \in \widetilde{\mathcal{N}}_P(v)} x_u^{\text{num}}
\end{equation}
where \(x_u^{\text{num}}\) denotes either a real-valued numerical feature or a categorical vector encoded as one-hot or multi-hot. This formulation allows us to summarize the distributional properties of both continuous and discrete structured features, providing a complementary signal to the textual summaries.

\begin{table*}[!t]
\centering
\small
\begin{tabular}{@{}ccccccccc@{}}
\toprule
Dataset & Node Type & Textual & Numerical & \# Nodes & \# Edges & \# Edge Types & \# Frauds & \# Train / Val / Test \\ \midrule
YelpReviews & Service Review & \cmark & \cmark & 67,395 & 17,486,608 & 3 & 8,919 & 1,348 / 1,348 / 13,479 \\
AmazonVideo & Product Review & \cmark & \cmark & 37,126 & 9,883,406 & 3 & 4,379 & 1,299 / 1,299 / 7,425 \\
E-Commerce & Shop Profile & \cmark & \cmark & 182,043 & 27,196,608 & 9 & 3,256 & 1,309 / 1,309 / 3,928 \\ 
LifeService & Shop Profile & \cmark & \cmark & 12,868 & 82,912 & 5 & 2,868 & 1,287 / 1,287 / 2,574 \\
\bottomrule
\end{tabular}
\caption{Overview of the datasets.}
\label{tab:dataset-stats}
\end{table*}

\subsection{Fraud Detection with DGP}

To incorporate textual and numerical features, we construct structured prompts:
\begin{equation}
\text{prompt}(v) = x^{\text{text}}_v \oplus \left[\bigoplus_{P \in \mathcal{P}} \big( S_P(v) \oplus a_P(v) \big) \right]
\end{equation}

We finetune the LLM on labeled nodes by minimizing the cross-entropy loss over the first generated token:
\begin{equation}
\mathcal{L} = -\frac{1}{|V_{\text{train}}|} \sum_{v \in V_{\text{train}}} \log p_\theta(y_v \mid \text{prompt}(v))
\end{equation}
where \(y_v \in \{\texttt{Yes}, \texttt{No}\}\) denotes the correct answer, and \(p_\theta(y_v \mid \text{prompt}(v))\) represents the token probability output by the LLM.

During inference, we apply a softmax function over the logits of the first generated token for the fraud probability:
\begin{equation}
p_v = \frac{\exp(\text{logit}_{\texttt{Yes}})}{\exp(\text{logit}_{\texttt{Yes}}) + \exp(\text{logit}_{\texttt{No}})}
\end{equation}
where \(\text{logit}_{\texttt{Yes}}\) and \(\text{logit}_{\texttt{No}}\) are the pre-softmax scores assigned by the model to the tokens \texttt{Yes} and \texttt{No}, respectively. The fraud probability \(p_v\) is interpreted as the model's confidence that node \(v\) is fraudulent.

\subsection{Complexity Analysis of DGP}

\paragraph{Token Consumption}
For a single target node, let $L$ denote the average token length of a node's text, $D$ the average out-degree, $R$ the number of relation types, $B$ the summarization budget, $K$ the number of hops, and $M$ the metapath neighbor truncation. 
The prompt length for a full-neighbor approach is $\frac{D^{K+1} - 1}{D - 1}L$. Similarly, a fully-vectorized prompt consumes $L + \frac{D^{K+1} - D}{D - 1}$ tokens, condensing neighbor texts into vectors. For DGP, the bi-level summarization prompts use $L + \frac{R^{(K+1)} - R}{R - 1}MB$ tokens, which scales with $R^K$ instead of $D^K$. The final prompt consumes $L + \frac{R^{(K+1)} - R}{R - 1}B$ tokens for fraud detection on the target node.

Since $R$, $K$, and $M$ are typically small in practice, DGP achieves significant token savings compared to full-neighbor methods. We note that $D$ can be much larger than $R$ in real-world heterogeneous graphs (\eg, $D = 133$ on the Amazon dataset), resulting in much longer full-neighbor prompts. Meanwhile, the average node text length $L$ continues to grow in modern web-scale datasets (\eg, $L = 170$ on the Yelp dataset), further amplifying the advantage of DGP’s bi-level summarization design.

\paragraph{Time Complexity}
The training phase consists of two frozen‑LLM summarization passes and one fine‑tuning loop. Processing all $N$ nodes with both node- and metapath-level summaries costs $\mathcal{O}\bigl((L+B)^2N\bigr)$ and $\mathcal{O}\bigl((\frac{R^{K+1} - R}{R - 1}MB)^2N\bigr)$, respectively. Finetuning on $\mathcal{O}(N)$ labeled nodes for $E$ epochs, each with sequence length $L + \frac{R^{K+1} - R}{R - 1}B$, costs $\mathcal{O}\bigl((L + \frac{R^{K+1} - R}{R - 1}B)^{2} EN\bigr)$. 

During inference, the two‑level summaries are cached, so each of the $N$ nodes requires only one forward pass of length $L + \frac{R^{K+1} - R}{R - 1}B$, giving $\mathcal{O}\bigl((L + \frac{R^{K+1} - R}{R - 1}B)^2N\bigr)$. 
Thus, the overall inference complexity scales linearly with the number of nodes $N$ and remains unaffected by large out-degree $D$, highlighting DGP's practicality in real-world applications. 
Importantly, the ratio between the target node's input length $L$ and the total neighbor input length $\frac{R^{K+1} - R}{R - 1} B$ can be flexibly controlled via the hyperparameters $K$ and $B$. This design mitigates the risk of neighbor information dominating the prompt and reduces computation on large multi-hop neighborhoods, ensuring that the model remains focused on the target node while maintaining practicality.

\begin{table*}[!t]
\centering
\resizebox{\textwidth}{!}{
\small
\begin{tabular}{@{}c|ccc|ccc|ccc|ccc@{}}
\toprule
Dataset & \multicolumn{3}{c|}{YelpReviews} & \multicolumn{3}{c|}{AmazonVideo} & \multicolumn{3}{c|}{E-Commerce} & \multicolumn{3}{c}{LifeService} \\
Method & MacroF1 & AUROC & AUPRC & MacroF1 & AUROC & AUPRC & MacroF1 & AUROC & AUPRC & MacroF1 & AUROC & AUPRC \\ \midrule
MLP & \ms{62.09}{0.05} & \ms{75.00}{0.06} & \ms{32.24}{0.13} & \ms{61.74}{0.39} & \ms{70.47}{0.18} & \ms{26.55}{0.48} & \ms{65.21}{0.08} & \ms{71.02}{0.11} & \ms{68.14}{0.20} & \ms{87.98}{0.08} & \ms{95.49}{0.04} & \ms{88.15}{0.21} \\
SAGE & \ms{64.64}{0.69} & \ms{75.59}{0.89} & \ms{36.75}{1.99} & \ms{62.23}{0.08} & \ms{70.11}{0.37} & \ms{25.93}{0.54} & \ms{68.66}{0.54} & \ms{73.85}{0.77} & \ms{71.84}{1.21} & \ms{88.42}{0.34} & \ms{95.42}{0.26} & \ms{88.09}{0.75} \\
HGT & \ms{66.53}{0.57} & \ms{81.49}{0.50} & \ms{40.04}{1.64} & {\ul \ms{65.55}{0.59}} & \ms{73.07}{0.70} & {\ul \ms{33.41}{0.66}} & \ms{65.07}{0.75} & \ms{72.05}{0.82} & \ms{68.42}{1.17} & \ms{89.28}{0.30} & \ms{95.82}{0.25} & \ms{89.90}{1.16} \\
ConsisGAD & {\ul \ms{67.33}{0.05}} & {\ul \ms{82.12}{0.21}} & {\ul \ms{42.11}{0.16}} & \ms{63.92}{1.89} & \ms{74.07}{1.83} & \ms{29.73}{2.42} & {\ul \ms{69.58}{0.50}} & {\ul \ms{77.10}{0.36}} & {\ul \ms{76.40}{0.40}} & {\ul \ms{90.55}{0.28}} & {\ul \ms{96.98}{0.12}} & {\ul \ms{92.85}{0.24}} \\
PMP & \ms{63.76}{2.87} & \ms{78.84}{0.71} & \ms{33.00}{1.15} & \ms{63.49}{3.10} & {\ul \ms{75.95}{0.99}} & \ms{30.40}{2.74} & \ms{66.44}{0.69} & \ms{74.47}{0.62} & \ms{70.25}{1.29} & \ms{88.41}{1.01} & \ms{95.95}{0.77} & \ms{89.62}{1.81} \\
GAAP & \ms{65.67}{3.14} & \ms{77.51}{3.63} & \ms{33.73}{1.30} & \ms{62.79}{2.28} & \ms{70.93}{2.84} & \ms{27.57}{2.51} & \ms{65.96}{0.70} & \ms{71.97}{0.66} & \ms{70.08}{0.70} & \ms{88.14}{0.63} & \ms{93.61}{0.92} & \ms{88.29}{0.76} \\ \midrule
LLM & \ms{60.79}{0.71} & \ms{71.18}{1.25} & \ms{30.45}{0.66} & \ms{59.90}{0.00} & \ms{71.78}{0.00} & \ms{27.03}{0.00} & \ms{63.87}{2.98} & \ms{67.97}{2.02} & \ms{66.57}{1.95} & \ms{89.19}{0.41} & \ms{96.40}{0.24} & \ms{91.32}{0.54} \\
TAPE & \ms{64.33}{1.81} & \ms{74.00}{1.41} & \ms{37.89}{2.03} & \ms{63.25}{1.61} & \ms{74.43}{1.81} & \ms{31.84}{1.64} & \ms{66.76}{0.83} & \ms{70.66}{1.10} & \ms{68.78}{1.70} & \ms{90.12}{1.29} & \ms{96.53}{1.86} & \ms{92.10}{1.37} \\
GraphGPT & \ms{60.96}{0.99} & \ms{70.39}{1.19} & \ms{30.66}{1.48} & \ms{59.13}{2.06} & \ms{70.76}{2.09} & \ms{27.82}{1.90} & \ms{64.12}{1.74} & \ms{67.38}{1.40} & \ms{67.85}{2.50} & \ms{87.54}{2.42} & \ms{91.62}{1.85} & \ms{87.59}{2.74} \\
HiGPT & \ms{62.49}{0.58} & \ms{71.80}{0.78} & \ms{31.59}{0.24} & \ms{60.99}{1.85} & \ms{71.49}{2.30} & \ms{29.69}{2.37} & \ms{66.70}{0.67} & \ms{69.80}{0.24} & \ms{67.62}{0.16} & \ms{89.61}{2.08} & \ms{96.15}{2.37} & \ms{90.15}{1.70} \\
InstructGLM & \ms{66.43}{0.14} & \ms{75.73}{0.57} & \ms{38.36}{0.24} & \ms{62.84}{0.21} & \ms{73.47}{0.43} & \ms{31.29}{0.47} & \ms{67.28}{1.19} & \ms{73.82}{1.58} & \ms{70.28}{2.23} & \ms{89.79}{1.57} & \ms{95.28}{1.31} & \ms{92.20}{2.16} \\ \midrule
DGP & \textbf{\ms{69.07}{0.23}} & \textbf{\ms{84.28}{0.11}} & \textbf{\ms{48.87}{0.82}} & \textbf{\ms{66.91}{0.13}} & \textbf{\ms{77.32}{0.11}} & \textbf{\ms{34.63}{0.24}} & \textbf{\ms{75.01}{0.11}} & \textbf{\ms{82.74}{0.28}} & \textbf{\ms{82.35}{0.20}} & \textbf{\ms{93.73}{0.26}} & \textbf{\ms{98.04}{0.06}} & \textbf{\ms{95.45}{0.07}} \\ \bottomrule
\end{tabular}
}
\caption{Comparison of fraud detection performance (\%) on different datasets.}
\label{tab:results}
\end{table*}

\subsection{Attention Dilution under Class Imbalance}
\label{sec:theory}

We provide theoretical insights into how excessive neighbor information overwhelm fraud signals, and demonstrate how our approach mitigates this issue. Let the input to the Transformer-based LLM backbone~\cite{vaswani2017attention} consist of $L$ tokens representing the target node and $m n_K$ tokens representing its $K$‑hop neighbors, where the total sequence length is $T_K = L + m n_K$. We denote the contribution from each neighbor node as \(m\) tokens, and the size of the $K$-hop neighborhood as $n_K = \frac{D^{K+1}-D}{D-1}$, which expands exponentially with the average out-degree $D$. We assume the target node is a fraudulent node we aim to detect. Suppose the global fraud ratio is $p \ll 1$, which is common in real-world graphs. As the number of neighbors increases, the fraction of fraud-related tokens in the prompt (including the target node itself) is given by:
\begin{equation}
r = \frac{L + p\,m n_K}{L + m n_K} = p + \frac{L(1 - p)}{L + m n_K} \le p + \frac{L(1 - p)}{m n_K}
\end{equation}
which gradually decreases from 1 to $p$. Notably, the softmax attention mechanism~\cite{vaswani2017attention} is given by:
\begin{equation}
\alpha_i = \frac{\exp(q^\top k_i/\sqrt{d})}{\sum_{j=1}^{T_K}\exp(q^\top k_j/\sqrt{d})},
\end{equation}
where $d$ denotes the model dimension, $q$ is the query vector (\ie, the last input token), and $k$ represents the key vectors. Thus, the cumulative softmax attention assigned to all fraud-related tokens is exactly $r$, which rapidly diminishes among the large number of benign tokens. This effect is well-studied as attention dispersion or over-squashing~\cite{liu2024lost,barbero2024transformers,vasylenko2025long}, where important signals are easily overwhelmed by irrelevant context in long sequences. 
Our bi-level summarization alleviates this issue by controlling $m$ and $n_K$, allowing the model to focus on informative fraud patterns.

\section{Experiments}
\label{sec:experiments}

We conduct extensive experiments to evaluate DGP from three perspectives: (i)~\textit{effectiveness} — how well DGP performs on fraud detection tasks over real-world heterogeneous graphs compared to GNN and LLM baselines; (ii)~\textit{component analysis} — how each design choice, such as textual and numerical summarization, affects performance; and (iii)~\textit{robustness} — how sensitive DGP is to hyperparameters and the design of summarization prompts.

\subsection{Experimental Setup}

\paragraph{Datasets}
We conduct experiments on four graph datasets, including two public benchmarks. 
YelpReviews~\cite{rayana2015collective} is a review-level spam detection dataset, where each node represents a review labeled as spam or non-spam. Following prior work~\cite{dou2020enhancing}, we construct a heterogeneous graph with three types of edges: reviews written by the same user (R-U-R), reviews on the same product with the same star rating (R-S-R), and reviews posted in the same month for the same product (R-T-R). Instead of using the handcrafted features introduced in the original work~\cite{rayana2015collective}, we directly utilize the original texts for LLM-based methods.
Amazon~\cite{mcauley2013amateurs} is a product review dataset from the Amazon Video category. We follow a similar graph construction, where each node is a review labeled as helpful or unhelpful. The graph contains three types of edges: reviews posted by the same user (R-U-R), reviews posted on the same product (R-P-R), and same-product reviews posted with the same rating and within the same week (R-S-R).

We also perform evaluation on two proprietary industry datasets: LifeService and E-Commerce, which are real-world graphs sampled from our industry partner, ByteDance.
Table~\ref{tab:dataset-stats} presents the key properties of the datasets. All datasets are characterized by mixed textual/numerical features, multi-type edges, and imbalanced fraud labels. Given the high cost of manual annotation in industrial settings, we construct training sets with a limited number of labeled samples, simulating realistic constraints where high-quality fraud labels are costly and difficult to obtain. 
We also note that the sum of the dataset split sizes, including the training, validation, and test sets, can be smaller than the total number of nodes. This aligns with a real-world scenario in which the majority of nodes are unlabeled, leaving them outside the regular data splits.

\paragraph{Baselines}
We benchmark against a wide range of competitive models: 
(i) \textit{GNNs}, including GraphSAGE~\cite{hamilton2017inductive}, HGT~\cite{hu2020heterogeneous}, ConsisGAD~\cite{chen2024consistency}, PMP~\cite{zhuo2024partitioning}, and GAAP~\cite{duan2025global};
(ii) \textit{Graph-agnostic models}, including MLP~\cite{rosenblatt1958perceptron} and a Qwen3-8B LLM~\cite{qwen3technicalreport} finetuned on target nodes alone; 
(iii) \textit{LLM-enhanced GNNs}, represented by TAPE~\cite{he2024harnessing};
and (d) \textit{graph-enhanced LLMs}, including GraphGPT~\cite{tang2024graphgpt}, HiGPT~\cite{tang2024higpt}, and InstructGLM~\cite{ye2023language}.
All baselines are implemented using official code.

\paragraph{Parameter Settings}

For all evaluated models, we tune hyperparameters using grid search based on validation performance. For DGP, we tune the bi-level summarization budgets $B_{\text{node}}, B_{\text{meta}} \in \{10, 20, 40, 80\}$, the the number of hops $K \in \{1,2,3\}$, and the neighbor truncation size $M \in \{2, 4, 8, 16\}$ for each dataset. 
For LoRA-based finetuning of LLM methods, we tune the LoRA rank $r \in \{4, 8, 16, 32\}$, the LoRA dropout rate $\in \{0.0, 0.05, 0.1\}$, and the learning rate $\in \{1\text{e}{-5}, 3\text{e}{-5}, 1\text{e}{-4}\}$. 
We set the batch size to 4 and finetune for up to 10 epochs with early stopping based on validation loss. 
For all baseline methods, we tune hyperparameters within the recommended ranges reported in their original papers to ensure fair and optimized comparisons.
All hyperparameters are selected to optimize the average AUROC on the validation set.

\paragraph{Implementation Details.}
We conduct all experiments on a machine with 4$\times$NVIDIA A100 GPUs (80GB). 
For all LLM-tuning methods, we use Qwen3-8B LLM backbone~\cite{qwen3technicalreport} for fair comparison. 
We insert LoRA~\cite{hu2022lora} adapters into all attention layers and use the AdamW~\cite{DBLP:conf/iclr/LoshchilovH19} optimizer for finetuning.
We adopt classification metrics including Macro-F1, AUROC, and AUPRC, and report the mean and standard deviation over 5 random seeds. 
All evaluation metrics are computed using the scikit-learn library~\cite{scikit-learn}.

\subsection{Performance Evaluation}

As shown in Table~\ref{tab:results}, DGP achieves consistent and substantial improvements over all baselines across datasets and evaluation metrics. We make the following observations:
\begin{itemize}
    \item GNN models generally outperform MLP, demonstrating the importance of leveraging graph structural information for fraud detection in heterogeneous graphs. Advanced GNNs such as ConsisGAD, which incorporate more sophisticated structural modeling, achieve better performance by capturing complex graph semantics.
    \item Although the datasets contain rich textual information, recent standalone LLMs (without graph context) generally underperform MLPs and fail to effectively leverage text for fraud detection.
    \item Methods like TAPE and InstructGLM combine graphs with LLMs, outperforming standalone LLMs by leveraging relational information. However, their performance remains constrained by challenges such as complex edge relations and neighborhoods in fraud detection scenarios, which can introduce noise and reduce effectiveness.
    \item Compared to encoding-based graph-enhanced LLMs, DGP avoids early vectorization and preserves more graph and textual information throughout the reasoning process. Compared to text-only LLMs, DGP uses dual-granularity semantic summarization to compress multi-hop neighborhoods, reducing neighbor domination and information overload. These design choices enable DGP to achieve the best overall results. Notably, DGP surpasses the strongest GNN baselines, suggesting that LLMs, when properly enhanced with graph context and summarization, hold significant potential for graph-based fraud detection.
\end{itemize}

\subsection{Detailed Analysis}

\subsubsection{Ablation Study}

\begin{figure}[!t]
  \centering
  \begin{subfigure}[b]{0.495\linewidth}
    \centering
    \includegraphics[width=\linewidth]{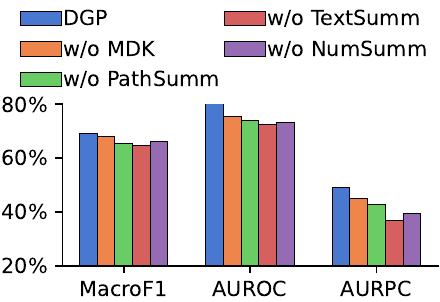}
    \caption{Yelp}
  \end{subfigure}
  \hfill
  \begin{subfigure}[b]{0.495\linewidth}
    \centering
    \includegraphics[width=\linewidth]{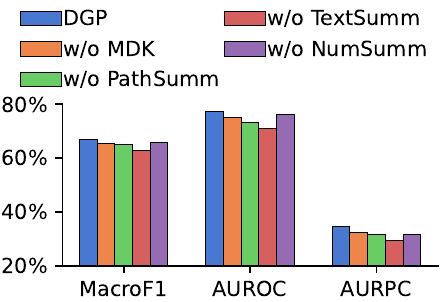}
    \caption{Amazon}
  \end{subfigure}
  \caption{Ablation study on DGP components. ``Path'' denotes \textit{Metapath}, ``Num'' denotes \textit{Numerical}, and ``Summ'' denotes \textit{Summarization}.}
  \label{fig:ab}
\end{figure}

To evaluate the effectiveness of each component in DGP, we conduct comprehensive ablation experiments on the Yelp and Amazon datasets. As shown in Figure~\ref{fig:ab}, we remove individual modules to obtain DGP variants and observe the resulting performance changes.

\begin{itemize}
\item Removing major components, including the textual summarization (w/o TextSumm) or numerical summarization (w/o NumSumm) modules, results in a clear performance drop. This demonstrates that these components are essential for capturing semantic and statistical signals in heterogeneous fraud graphs. Specifically, textual features exhibit greater importance than numerical features, indicating that text in these datasets is particularly informative for fraud detection.
\item We also ablate subsidiary components within textual summarization, including Markov Diffusion Kernel-based metapath trimming (w/o MDK) and metapath summarization (w/o PathSumm). Removing either component leads to a performance decline, indicating their positive contributions to generating informative representations of neighbor nodes.
\end{itemize}

\subsubsection{Impact of Summarization Length}

\begin{figure}[!t]
  \centering
  \begin{subfigure}[b]{0.495\linewidth}
    \centering
    \includegraphics[width=\linewidth]{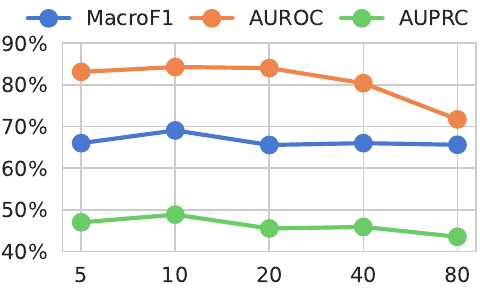}
    \caption{Yelp}
  \end{subfigure}
  \hfill
  \begin{subfigure}[b]{0.495\linewidth}
    \centering
    \includegraphics[width=\linewidth]{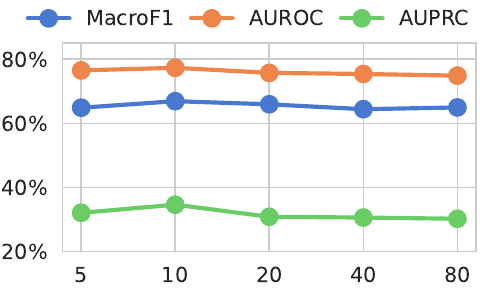}
    \caption{Amazon}
  \end{subfigure}
  \caption{Impact of summarization length (words).}
  \label{fig:ab2}
\end{figure}

We further examine the effect of varying the neighbor summary token budget $B$. Figure~\ref{fig:ab2} presents fraud detection metrics across a range of budgets $B \in \{5, 10, 20, 40, 80\}$ for the YelpReviews and AmazonVideo datasets. For simplicity, we assume a unified budget, \ie, $B_{\text{node}} = B_{\text{meta}}$. We observe that very short summaries (5 tokens) provide insufficient context and thus degrade performance, while longer summaries (\eg, 80 tokens) may lead to token dilution, also reducing accuracy. 

Notably, the best performance is generally achieved with a relatively small summarization budget (10 tokens). This suggests that highly coarse-grained summarizations of neighbor information are sufficient to enhance fraud detection on graphs. Importantly, this indicates that DGP remains effective in complex real-world graphs, as it does not rely on fine-grained or verbose descriptions of each neighbor.

\begin{table}[!t]
\centering
\small
\begin{tabular}{@{}ccccc@{}}
\toprule
Dataset & Task-Aware & Macro F1 & AUROC & AURPC \\ \midrule
\multirow{2}{*}{Yelp} & \xmark & \ms{69.07}{0.23} & \ms{84.28}{0.11} & \ms{48.87}{0.82} \\
 & \cmark & \ms{58.65}{0.05} & \ms{70.65}{1.05} & \ms{29.02}{0.36} \\ \midrule
\multirow{2}{*}{Amazon} & \xmark & \ms{66.91}{0.13} & \ms{77.32}{0.11} & \ms{34.63}{0.24} \\
 & \cmark & \ms{65.55}{0.21} & \ms{73.73}{0.17} & \ms{31.82}{0.27} \\ \bottomrule
\end{tabular}
\caption{Impact of node-level summarization prompts.}
\label{tab:fraud-aware}
\end{table}

\subsubsection{Impact of Task-Aware Summarization}

We analyze whether explicitly introducing fraud-aware summarization prompts helps or hinders DGP’s classification performance. Table~\ref{tab:fraud-aware} reports results for both task-agnostic and task-aware neighbor summarization strategies. In the task-agnostic setting, we use a generic instruction such as \texttt{Summarize the text within 10 tokens}. In contrast, the task-aware setting introduces domain-specific cues, \eg, \texttt{Summarize the text within 10 tokens, focusing on signals indicative of fraudulent behavior}.

The results suggest that task-aware summarization degrades DGP's performance. This may be due to reduced generality, where overly specific prompts constrain the LLM’s ability to capture subtle fraud signals. In contrast, task-agnostic prompts allow for broader cue discovery, potentially supporting more robust classification.

\section{Conclusion}

We introduced DGP, a framework for fraud detection on heterogeneous graphs that combines semantic-aware summarization, diffusion-based neighbor selection, and type-specific feature aggregation. By condensing relevant multi-hop textual contexts and precisely aggregating structured attributes, DGP enables effective fraud prediction using large language models. Extensive experiments demonstrate its superior performance and robustness across diverse benchmarks. In future work, we will explore dynamic graphs in which fraud patterns evolve over time.

\bibliography{aaai2026}

\clearpage
\newpage

\end{document}